\documentclass[sigconf]{acmart}
\AtBeginDocument{%
  \providecommand\BibTeX{{%
    \normalfont B\kern-0.5em{\scshape i\kern-0.25em b}\kern-0.8em\TeX}}}

\usepackage{amsmath,amsthm}

\usepackage{latexsym}
\usepackage{url}
\usepackage{makecell}
\usepackage{tablefootnote}
\usepackage{balance}
\usepackage[utf8]{inputenc}
\usepackage{bbding}
\usepackage{pifont}
\usepackage{wasysym}
\usepackage{graphicx}
\usepackage{subfig}
\usepackage{booktabs}
\usepackage{algorithm}
\usepackage{hyperref}
\usepackage{varwidth}
\usepackage{multirow}
\usepackage[export]{adjustbox}
\usepackage{algorithmic}
\copyrightyear{2023}
\acmYear{2023}
\setcopyright{licensedothergov}\acmConference[CIKM '23]{Proceedings of the 32nd ACM International Conference on Information and Knowledge Management}{October 21--25, 2023}{Birmingham, United Kingdom}
\acmBooktitle{Proceedings of the 32nd ACM International Conference on Information and Knowledge Management (CIKM '23), October 21--25, 2023, Birmingham, United Kingdom}
\acmPrice{15.00}
\acmDOI{10.1145/3583780.3614870}
\acmISBN{979-8-4007-0124-5/23/10}

\begin{document}

\title[Towards Knowledge Infused Multi-modal Clinical Conversation Summarization]{{\em Experience and Evidence are the eyes of an excellent summarizer!} Towards Knowledge Infused Multi-modal Clinical Conversation Summarization}

\author{Abhisek Tiwari}
\affiliation{%
  \institution{Indian Institute of Technology, Patna}
  \city{Patna}
  \country{India}}
\email{abhisek\_1921cs16@iitp.ac.in}
\author{Anisha Saha}
\affiliation{%
  \institution{Indian Institute of Technology, Patna}
  \city{Patna}
  \country{India}}
\email{anisha0325@gmail.com}

\author{Sriparna Saha}
\affiliation{%
\institution{Indian Institute of Technology, Patna}
  \city{Patna}
  \country{India}}
\email{sriparna@iitp.ac.in}

\author{Pushpak Bhattacharyya}
\affiliation{%
  \institution{Indian Institute of Technology, Bombay}
  \city{Bombay}
  \country{India}}
  \email{pb@cse.iitb.ac.in}

\author{Minakshi Dhar}
\affiliation{%
 \institution{All India Institute of Medical Sciences, Rishikesh}
 \city{Rishikesh}
 \country{India}}
 \email{minakshi.med@aiimsrishikesh.edu.in}



\begin{abstract}
 With the advancement of telemedicine, both researchers and medical practitioners are working hand-in-hand to develop various techniques to automate various medical operations, such as diagnosis report generation. In this paper, we first present a multi-modal clinical conversation summary generation task that takes a clinician-patient interaction (both textual and visual information) and generates a succinct synopsis of the conversation. We propose a knowledge-infused, multi-modal, multi-tasking medical domain identification and clinical conversation summary generation ({\em MM-CliConSummation}) framework. It leverages an adapter to infuse knowledge and visual features and unify the fused feature vector using a gated mechanism. Furthermore, we developed a multi-modal, multi-intent clinical conversation summarization corpus annotated with intent, symptom, and summary. The extensive set of experiments, both quantitatively and qualitatively, led to the following findings: (a) critical significance of visuals, (b) more precise and medical entity preserving summary with additional knowledge infusion, and (c) a correlation between medical department identification and clinical synopsis generation. Furthermore, the dataset and source code are available at \url{https://github.com/NLP-RL/MM-CliConSummation}.
\end{abstract}



\maketitle

\section{Introduction}
In the past few years, tele-health has grown immensely with the advancement of information \& communication technologies (ICTs) and artificial intelligence-based applications for healthcare activities \cite{nittari2020telemedicine}. With the COVID-19 pandemic, internet utilization for healthcare activities has reached its peak in the last two decades and has become a new normal \cite{wosik2020telehealth}. On the other hand, many recent healthcare surveys and the World Health Organisation (WHO) found an uneven doctor-to-population ratio, estimating a deficit of more than 12 million healthcare workers by 2030. Thus, tele-health usage is being actively encouraged by healthcare providers, and patients are adopting it at the same pace \cite{valizadeh2022ai}. One such manifestation that has become popular in both research and industry communities is automatic disease diagnosis (ADD) \cite{valizadeh2022ai}. ADD aims to assist doctor by conducting primary symptom and sign investigations, allowing them to focus on diagnosis and treatment. A few hospitals have already implemented diagnosis assistants like Ada\footnote{\url{https://ada.com/}}, and Mayo Clinic\footnote{\url{https://www.mayoclinic.org/}} for clinical assistance. 

\begin{figure*}
    \centering
    \includegraphics[scale=0.55]{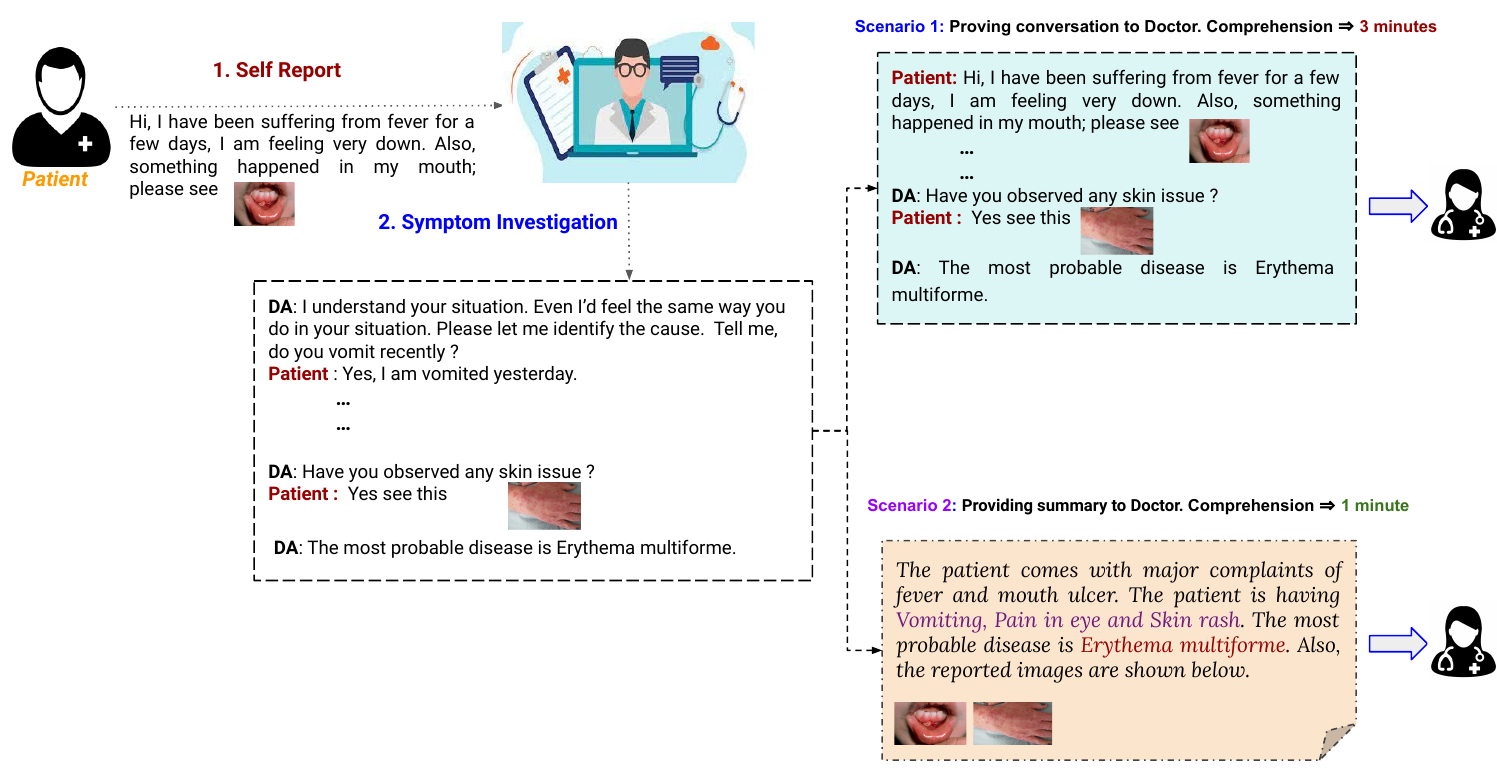}
    \vspace{-1.6em}
    \caption{An illustration of an autonomous symptom investigation and disease diagnosis assistant with and without conversation summary generation. The second scenario is evidently more comprehensible and time-efficient.}
    \label{Fig1_Prob}
\end{figure*}

When we consult with doctors, they often conduct a preliminary investigation by analyzing the patient's self-report and investigating other pertinent symptoms and signs. Even though, they may require some lab reports to confirm a medical condition, the initial investigation helps to decide on some lab tests. With this motivation, \citet{wei2018task} formulated a conversational artificial intelligence-based symptom investigation and diagnosis assistant. In our conversations, we often show our visual medical conditions, such as skin rash, to doctors for precise diagnosis and accurate treatment. Driven by the motivation, \citet{tiwari2022dr} first proposed a multi-modal automatic disease diagnosis virtual assistant called MDD-VA, which demonstrated the critical impact of considering the visual form of symptom reporting on diagnosis efficacy and end-user satisfaction. The diagnosis dialogue framework is illustrated in Figure \ref{Fig1_Prob} (left side). Certainly, it aids doctors by automatically collecting primary investigations. However, the diagnosis assistant forwards the entire conversation to doctor, which takes a significant amount of time to comprehend the same (Figure \ref{Fig1_Prob}, right top). On the other hand (Figure \ref{Fig1_Prob}, right bottom), a summary of the dialogue has been provided, which took significantly less time to comprehend the case. Furthermore, the summary could be utilized for storing the cases efficiently and thus enhancing its re-usability.

It is often stated that an image conveys an idea more effectively than thousand words. In today's digital media era, we are compelled to use images and visuals in our daily lives. We frequently rely on visuals in conversations, especially in clinical discussions, where we aim to convey medical conditions accurately. A recent survey conducted by \citet{popov2022industry} highlighted the substantial market value of medical imaging, which reached 32 billion in 2022 and is projected to grow at a compound annual growth rate of 4.9\%. This clearly underscores the importance of images and visuals in clinical dialogue settings. In real-life scenarios, if two individuals without medical expertise were asked to summarize two medical dialogues, the older individual would likely perform better. The key differentiating factor is the additional knowledge possessed by the older person. A corpus serves as a representation of behavior, and therefore, learning from a corpus, along with additional relevant knowledge, incorporates a global perspective \cite{tiwari2022knowledge}. Hence, inspired by the effectiveness of multi-modality and external knowledge, we explore fundamental research questions related to multimodal dialogue summarization and propose a novel transformer-based adapter-driven clinical conversation summarization framework.

\hspace{-0.35cm}\textbf{Research Questions} We aim to investigate the following three research questions related to multimodal clinical conversation summary generation in the paper: \textbf{(i)} How does the inclusion of visual cues, such as visual signs and patients' expressions, impact the process of clinical patient-doctor interaction summarization? \textbf{(ii)} Can the inclusion of external knowledge offer more relevant context, thereby enhancing the quality of generated medical dialogue summaries? Does the fusion mechanism of visual/knowledge information with text have any influence on the overall quality of the summary?  \textbf{(iii)} Is there a correlation between the identification of medical departments and the summarization of medical dialogues?

In order to build a multi-modal clinical conversation summary generation model and validate the research questions, we take the first attempt to build a multi-modal clinical conversation summary ({\em MM-CliConSumm}) dataset. The dataset bridges the following gaps: (i) Textual-Visual aided clinical interactions annotated summary. (ii) Each patient utterance is annotated with the medical entities contained in it and each dialogue with the concerned medical department and disease. (iii) We have also provided two additional executive summaries for interactions: (a) Medical Concern Summary (MCS), which is a concise one-line summary that captures the primary concern expressed by the patient during the discussion, (b) Doctor Impression (DI) encapsulates the final reaction and impression of the doctor following the conversation with the patient. \\
\hspace{-0.35cm}\textbf{Key Contributions} The key contributions of the work are fourfold, which are enumerated below. 
\vspace{-1em}
\begin{itemize}
    \item Motivated by the tremendous efficacy of visuals in clinical conversation settings, we first propose an autonomous task of multi-modal clinical conversation summarization (MM-CCS) and medical concern summary (MCS) generation.  
    \item We first created a multimodal medical conversation summarization dataset, named {\em MM-CliConSumm}, which contains clinical conversation annotated with medical vitals such as medical department, patient side summary, one-line summary, and overall summary.
    \item  We propose a multitasking knowledge-infused medical department identification and multi-modal clinical conversation summary generation ({\em MM CliConSummation}) model incorporated with an adapter-based contextualized M-modality fusion mechanism that evaluates visual abnormalities and infuses additional knowledge in conjunction with patient-doctor interaction.
    \item The proposed {\em MM CliConSummation} model outperforms existing state-of-the-art uni-modal medical summarization models and baselines across all evaluation metrics, including human evaluation.
\end{itemize}

\vspace{-0.5em}

\section{Related Work} The proposed work is relevant to the following three research areas: dialogue summarization, multi-modal summarization, and knowledge-infused text generation. In the following paragraphs, we have summarized the relevant works.

\hspace{-0.36cm}\textbf{Dialogue Summarization} Dialogue summarization has been a longstanding and fundamental problem in Natural Language Processing (NLP). Over the past two decades, the field of dialogue summarization has progressed in the following directions \cite{el2021automatic}: (i) Feature guided Extractive Summarization \cite{binwahlan2009swarm}, (ii) RNN-based summary generation \cite{lin2018global}, (iii) Pre-trained large language model (PLM) based summarization \cite{zhong2022dialoglm}. In the last few years, the focus has been on the aspect (domain/intent/keyword) guided dialogue summarization and, synthetic data creation with few shot settings. In  \cite{joshi2020dr}, the authors have proposed a summarization model based on a pointer network generator. The model takes dialogues as input and generates a summary for each turn (doctor-patient) of the interaction. The work \cite{song2020summarizing} proposed a hierarchical encoder-tagger for summarizing medical patient-doctor conversations by identifying important utterances.

\hspace{-0.36cm}\textbf{Multi-modal Summarization} Multi-modal summarization aims to generate coherent and important information from data having multiple modalities \cite{zhu2018msmo}. In the last few years, the main focus of multi-modal summarization has been to find co-relation among different modalities: text, audio, and image for video data \cite{apostolidis2021video}. An important segment of a video is a subjective concern and may also vary among consumers. In \cite{huang2021gpt2mvs}, the authors have proposed a new task of user constraint-based summarization and proposed an attention mechanism to summarize the query-relevant content. To generate a coherent summary, synchronizing different modalities is crucial. Shang et al., \cite{shang2021multimodal} proposed a time-aware multi-modal transformer (TAMT) that leverages time stamps across image, text, and audio to generate an adequate and coherent video summary.

\hspace{-0.36cm}\textbf{Knowledge-Infused Text Generation and Summarization}\\ Knowledge-infused text generation \cite{liu2022knowledge} incorporates external knowledge or information into the process of generating text, allowing the model to generate more accurate and relevant content \cite{gaur2020knowledge}. In \cite{wang2020improving}, the authors have proposed a novel knowledge-infused dialogue generation model that infuses additional knowledge provided by ConceptNet \cite{speer2017conceptnet} for query-type utterances, with dialogue context, demonstrating improved generation quality over traditional models. In dialogue, all utterances are not equally important. Motivated by the observation, Manas et al., (2021) \cite{manas2021knowledge} proposed PHQ-9 lexicon-guided clinical text-based conversation. They showed their proposed unsupervised model that infuses the knowledge and performs superior in terms of informativeness and underlying interview theme. However, the summaries produced are primarily template-driven and consist of a compilation of turn-specific synopses, resulting in a tedious and lengthy summary.
\vspace{-0.5em}

\section{Dataset} We first extensively scrutinized the existing benchmark clinical conversational datasets, and the summary is presented in Table \ref{ED}. We found Vis-MDD \cite{tiwari2022dr} as the most relevant dataset for the proposed multi-modal clinical conversation summarization task. Motivated by the unavailability and the efficacy of clinical conversation summary, we first take the move to develop a multimodal clinical conversation summary generation ({\em MM-CliConSumm}) dataset. We curated the dataset based on the Vis-MDD corpus under the guidance of two medical professionals.  

\begin{table*}[hbt!]
    \scalebox{0.72}{
    \begin{tabular}{lccccccccc}
    \hline
      \textbf{Dataset} & \textbf{Language}  & \textbf{Conversation} & \textbf{Image} & \makecell[c]{ \textbf{Sign's} \\ \textbf{Severity}}  &   \makecell[c]{ \textbf{Intent \&} \\ \textbf{Symptom}} & \makecell[c]{ \textbf{Medical} \\ \textbf{Department}} & \textbf{Summary} &   \makecell[c]{ \textbf{Patient Concern} \\ \textbf{Summary}} & \makecell[c]{ \textbf{Doctor} \\ \textbf{Impression}} \\ \hline
     RD \cite{wei2018task}  & Chinese & $\times$ & $\times$ & $\times$ & $\times$ & $\checkmark$ &  $\times$  & $\times$ & $\times$ \\ 
     DX \cite{xu2019end} & Chinese  & $\times$ & $\times$ &  $\times$ & $\times$ & $\checkmark$ &  $\times$ & $\times$ & $\times$\\ 
     M$^2$ \cite{yan2021m} & Chinese & $\checkmark$ & $\times$ & $\times$ & $\checkmark$ & $\checkmark$ &  $\times$ &  $\times$ & $\times$ \\
     MedDialog-EN \cite{zeng2020meddialog} & English & $\checkmark$ & $\times$ & $\times$ & $\times$ & $\times$ & $\times$ & $\times$ & $\times $  \\ 
     SD \cite{liao2020task} & English & $\times$ & $\times$ & $\times$ & $\times$ & $\checkmark$ &  $\times$ & $\times$ & $\times$\\ 
      Dr. Summarize \cite{joshi2020dr}  & English &  $\checkmark$ & $\times$ & $\times$ & $\times$ & $\times$ & $\times$ & $\times$ & $\times$ \\
      GPT3-ENS SS \cite{chintagunta2021medically} & English & $\checkmark$ & $\times$ & $\times$ & $\times$ & $\times$ & $\checkmark$ & $\times$ & $\times$ \\
      Vis-MDD \cite{tiwari2022dr} & English & $\checkmark$ & $\checkmark$ &  $\times$ & $\checkmark$ & $\checkmark$ &  $\times$ &  $\times$ & $\times$ \\
      \textbf{MM-CliConSumm}& English & $\checkmark$ & $\checkmark$ &  $\checkmark$ & $\checkmark$ & $\checkmark$ &  $\checkmark$ &  $\checkmark$ & $\checkmark$ 
      \\ \hline 
    \end{tabular}}
    \caption{Statistics of the existing publicly available medical datasets for disease diagnosis task}
    \vspace{-1em}
    \label{ED}
\end{table*}
 
\subsection{{\em MM-CliConSumm}}
We, along with the two medical experts, first analyzed a few medical dialogues of Vis-MDD dataset. We have provided a subset of 100 dialogue samples across ten medical departments, each having 10 samples to the clinicians for summary writing. It contained both text and image-based utterances in each conversation. They wrote three different kinds of summaries for each interaction: overall summary, medical concern summary/MCS (patient side short summary), and doctor impression (DI). The objective of annotating two new kinds of summaries was inspired by telemedicine. MCS helps online healthcare users to locate relevant information effectively, whereas doctor impression aims to help doctors and healthcare systems for effective reference and action points. We further asked them for annotation guidelines for summary writing and provided the sample dataset to three annotators (biology graduate students) for scaling up the corpus. In order to ensure annotation agreement among the annotators, we calculated the kappa coefficient (k). It was found to be 0.73, indicating a significant uniform annotation. The {\em MM-CliConSumm} dataset statistics are provided in Table \ref{DStat}.  

\begin{table}[hbt!]
    \centering
    \scalebox{0.77}{
    \begin{tabular}{lp{4.5cm}}
    \hline
       \textbf{Entries} & \textbf{Value} \\
       \hline 
       \# number of conversations   & 1668 \\
        \# of utterances &  5483 \\
       \# of unique words & 3512\\ 
       \# of unique images & 1668 \\
       \# number of symptoms & 266 \\
        \# number of diseases & 90 \\
        \# number of medical departments & 10 \\
        \# of diseases in each department & 9 \\
       avg. length of overall summary (\# of words) & 48 \\
       avg. length of MCS (\# of words) & 16.64 \\
       avg. length of Doctor impression (\# of words) & 16.86 \\
       tags & intent, symptom, visual information, overall summary, MCS, and doctor impression \\
       \hline
    \end{tabular}}
    \caption{{\em MM-CliConSumm} dataset statistics}
    \vspace{-2em}
    \label{DStat}
\end{table}

\vspace{-0.7em}
\subsection{Qualitative Aspects} The objective of summarization is to represent the essence of a large document in a precise yet concise manner without losing any critical characteristics. To generate an adequate summary of a clinical conversation, we analyze different qualitative characteristics of clinical interactions and incorporate them accordingly.

\begin{figure}[hbt!]
    \includegraphics[scale=0.39]{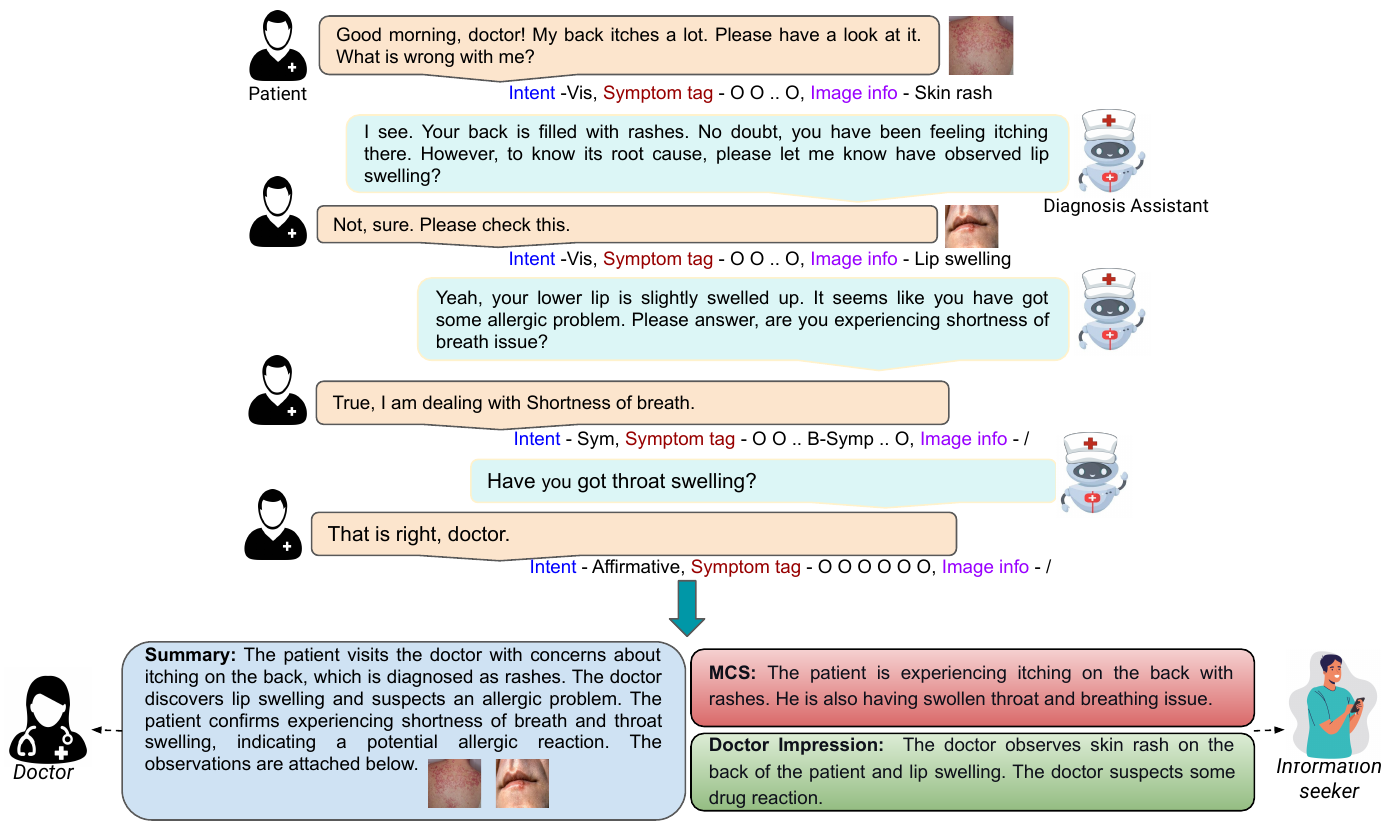}
    \vspace{-2.25em}
    \caption{A data sample from the {\em MM-CliConSumm} corpus}
    \label{DSS}
\end{figure}

\hspace{-0.36cm}\textbf{Importance of Visual Descriptions} The medical domain is highly specialized and sensitive, and many individuals are unfamiliar with various medical terms, such as "mouth ulcer" and "skin growth". Furthermore, we make an effort to communicate our medical conditions as accurately as possible. Describing something like a skin rash and its intensity through text can be challenging (as shown in Figure \ref{DSS}), so presenting the actual medical condition visually offers an easier and more precise means of communication.

\begin{figure}[hbt!]
    \centering
    \includegraphics[scale=0.44]{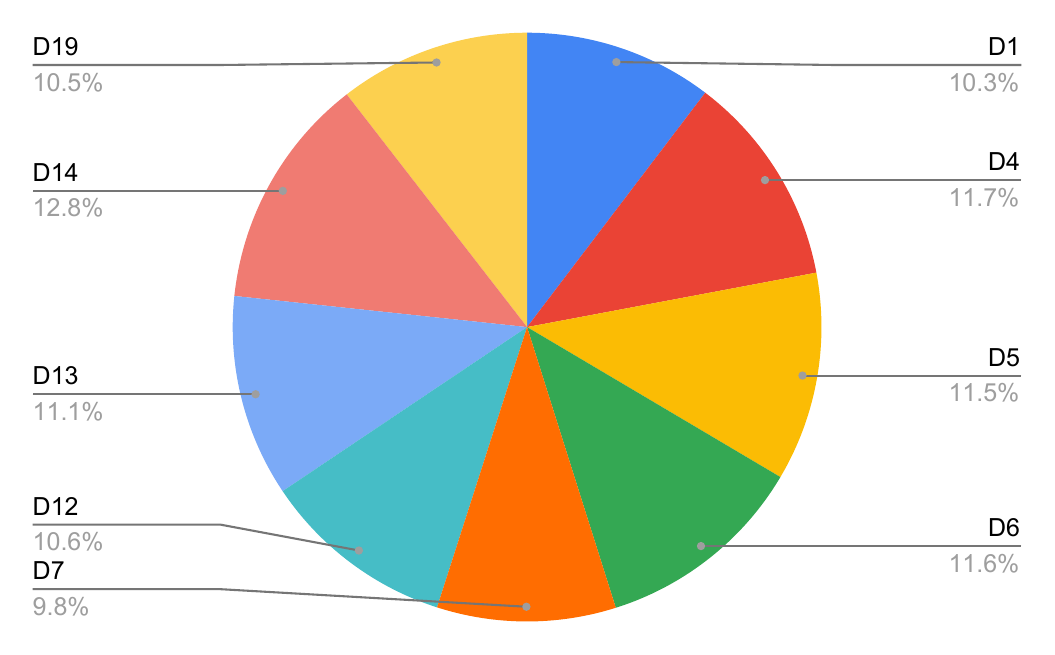}
    \vspace{-1.7em}
    \caption{Distribution of conversations across different medical departments in the corpus}
    \label{MDD}
\end{figure}

\hspace{-0.36cm}\textbf{Importance of Medical Department}
Labeling clinical conversations with medical departments can be beneficial for both clinicians and online healthcare users, as it allows for easy referencing and reusability. Moreover, the medical department label can assist in generating domain-guided responses and summaries of the conversation, ensuring that the information provided is more relevant and tailored to the specific medical field. The distribution of different medical departments in the curated dataset is provided in Figure \ref{MDD}. There are 9 medical departments, and each group contains 10 different diseases. The division is determined as per International Classification
of Diseases (ICD-10-CM). 


\hspace{-0.36cm}\textbf{Role of Medical Concern Summary (MCS)} The Medical Concern Summary (MCS) is a concise and focused summary of a patient's main concern, which is discussed during the interaction with a clinician. Its purpose is to assist online healthcare users in quickly identifying whether a clinical conversation contains the information they are seeking. As an example, the MCS presented in Figure \ref{DSS} encapsulates the essence of the entire conversation, allowing users to easily determine whether they should refer to the content or not.

\hspace{-0.36cm}\textbf{Importance of Doctor Impression (DI)}
In the process of clinical diagnosis and treatment, a patient's journey typically involves multiple interactions rather than a single visit. Consequently, reviewing the entire transcript of a previous lengthy conversation can be time-consuming. Therefore, having access to the patient's Medical Concern Summary (MCS) along with the doctor's impression (as shown in Figure \ref{DSS}) serves as a helpful synopsis/action points of the case for different healthcare stakeholders, reducing the need to refer to the lengthy transcript.

\hspace{-0.36cm}\textbf{Ethical Consideration} We strictly followed the medical research's legal, ethical, and regulatory guidelines during the dataset curation process. With this in mind, we have not added or removed any utterances from the conversation. The curated dataset does not reveal users' identities, such as their names and demographic information. The annotation guidelines are provided by the clinicians, and the dataset is thoroughly checked and corrected by them. Furthermore, we have also obtained approval from our institute's healthcare committee and institutional ethical review board (ERB) to employ the dataset and carry out the research. 

\section{Proposed Methodology}
\begin{figure*}[hbt!]
    \centering
    \includegraphics[scale=0.71]{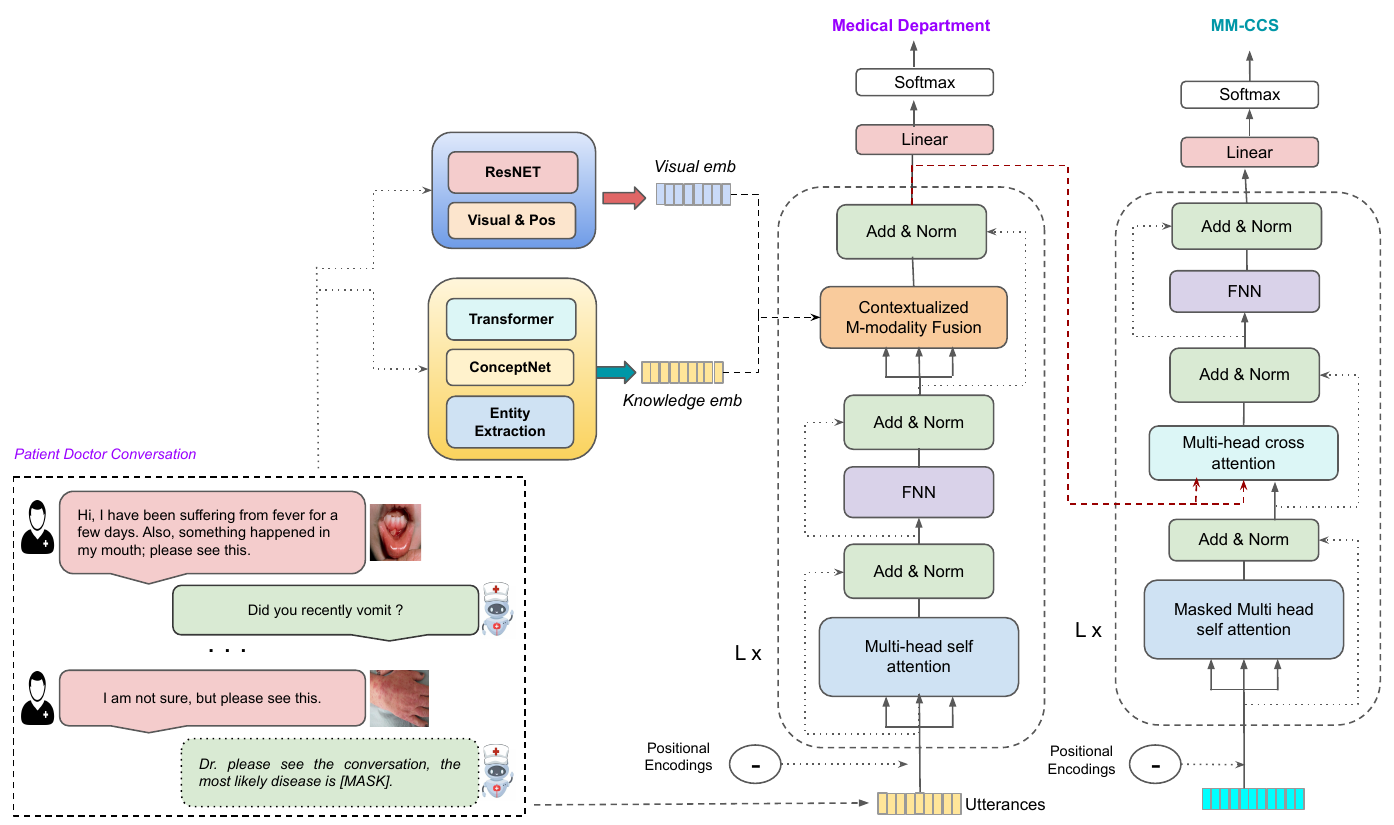}
    \vspace{-1em}
    \caption{Architecture of the proposed multi-tasking, multimodal medical department identification, and summary generation ({\em MM-CliConSummation}) model}
    \label{PM}
\end{figure*}

We anticipate that multi-modal clinical interaction summarization has crucial importance of the following in addition to text/speech of clinicians and patients: (a) patient's visual reporting during an interaction, (b) additional relevant knowledge, and (c) concerned medical department. Thus, we propose a multi-tasking, multi-modal, knowledge-infused medical department identification and dialogue summary generation framework. The proposed architecture is illustrated in Figure \ref{PM}. We introduce the novel concept of Contextualized M-modality fusion, which utilizes an adapter-based module in a transformer to effectively integrate order-driven visuals and external relevant knowledge for dialogue summarization. There are three key stages: (i) Discourse, Visual and Knowledge representation, (ii) Contextualized M-modality fusion, and (iii) Clinical department identification and Summary generation. The working of each stage and the involved module is explained and illustrated in the subsequent sections.
\subsection{Discourse, Visual and Knowledge Representation} We have employed three types of information to encode a patient-clinician interaction: discourse text (utterances from both the clinician and the patient), visuals that were discussed during the interaction, and supplementary discourse-related information obtained from general knowledge. The process of encoding the entities is described below.

\hspace{-0.36cm}\textbf{Discourse Representation} A discourse consists of a sequence of utterances from both the patient and the doctor, where the patient explains his/her main concerns and the doctor conducts further inquiries to aid in diagnosis. We tokenize the combined patient-doctor texts, segmented by turns, using BART tokenizers \cite{lewis2019bart}, and extract the transcript embedding.

\hspace{-0.36cm}\textbf{Visual Features} To leverage the pre-training of a state-of-the-art model, we opted for ResNet 152 \cite{he2016deep}, a widely used visual model, to represent the images. We began by fine-tuning the ResNet model using our labeled dataset of 1700 images, each with corresponding symptom or sign labels. For the symptom identification task, we added a neural network on top of the ResNet and froze the weights of all layers except the last three. Eventually, we extracted the vector representation of the image by pooling the output of the last layer of the ResNet. In cases where multiple images were present within a single interaction, we computed the mean of the image embeddings for visual representation.

\hspace{-0.36cm}\textbf{Knowledge Infusion} The selection of content to include in a summary is a crucial aspect of summarization. This decision-making process is influenced by factors such as the number of samples in the dataset and their diversity. However, in the medical domain, the dataset size is not extremely large due to its sensitivity. In such cases, prior experience and relevant additional knowledge can be particularly valuable. Hence, we utilize additional knowledge to assist the generation model in emphasizing pertinent content. 

\begin{algorithm}[hbt!]
\algsetup{linenosize=\tiny}
 \scriptsize
\caption{Discourse aware Knowledge Distillation (DKD)}
\textbf{Input} Context ($C: p_1, d_1, p_2, d_2,... d_n$) where $p_i$ and $d_i$ represents $i^{th}$ utterances of patient and doctor, respectively \\
\hspace{-3.95cm}\textbf{Output} Context relevant Knowledge Graph ($KG_C$) \\
\hspace{0.03cm}\textbf{Initialization} $n_k$ (7): threshold for the number of keywords from a conversation, $n_r$ (5): threshold for the number of concepts for an entity
\begin{algorithmic}[1] 
\STATE  $KG_C$ = []  
\STATE K[$1, 2, .... n_k$] = YAKE($C, n_k$) \hfill$\Rightarrow$  $K$: list of entities
\FOR{entity in K}
\STATE $KGT_{entity}$ =[] \hfill$\Rightarrow$  $KGT_{entity}$: KG triplet for ``entity'' 
\FOR{j in range(0, $n_r$)}
\STATE  <$r_j, h_j, t_j$> = ConceptNet(entity, $KGT_{entity}$)  \hfill$\Rightarrow$  r: relation, h: head and t: tail
\STATE $KGT_{entity}$ = $KGT_{entity}$ + [$r_j, h_j, t_j$]
\ENDFOR
\STATE $KG_C$ = $KG_C$ + $KGT_{entity}$
\ENDFOR
\STATE \textbf{return} $KG_C$ 
\end{algorithmic}
\label{algo}
\end{algorithm} 

We choose ConceptNet \cite{speer2017conceptnet} for knowledge infusion, which is one of the largest knowledge graphs (8 million nodes and 21 million edges) that contains concepts of various domains, such as healthcare. While knowledge is crucial, focusing on relevant knowledge is more significant while solving a task. Thus, infusing the entire ConceptNet knowledge with the proposed summarization setup would be ineffective and may even deteriorate the performance because a large chunk of it would be irrelevant in a very large number of cases. We propose to distill the external knowledge based on discourse and inject a subset of the knowledge graph dynamically depending on the context. It first extracts essential words (keywords) from the dialogue using an unsupervised statistical-based keyword extractor called YAKE \cite{campos2020yake}. The extracted entities are passed to the ConceptNet, which identifies relevant concepts associated with them as described in the Algorithm \ref{algo}.

\subsection{Contextualized M-modality Fusion} The manner in which multiple pieces of information are integrated together holds substantial importance for the effectiveness of the combined representation. Therefore, it is vital to merge them in a manner that transforms them into a unified embedding space, ensuring the coherence of the combined representation. Motivated by this, we propose an adapter-based infusion mechanism called contextualized M-modality fusion for combining text, image, and knowledge, which is effective to incorporate with transformer models. The contextualized M-modality generates contextualized modality-conditioned key and value vectors and produces a scaled dot product attention vector. The contextualized modality attention vector is being utilized for calculating the global information attended over visual and knowledge information, which is being utilized for medical domain identification and clinical summary generation. The infusion mechanism is illustrated in Figure \ref{IM}. It takes the hidden state (H) and calculates the contextualized modality attention as follows:
\vspace{-1.25em}
\begin{figure}[hbt!]
    \centering
    \includegraphics[scale=0.7]{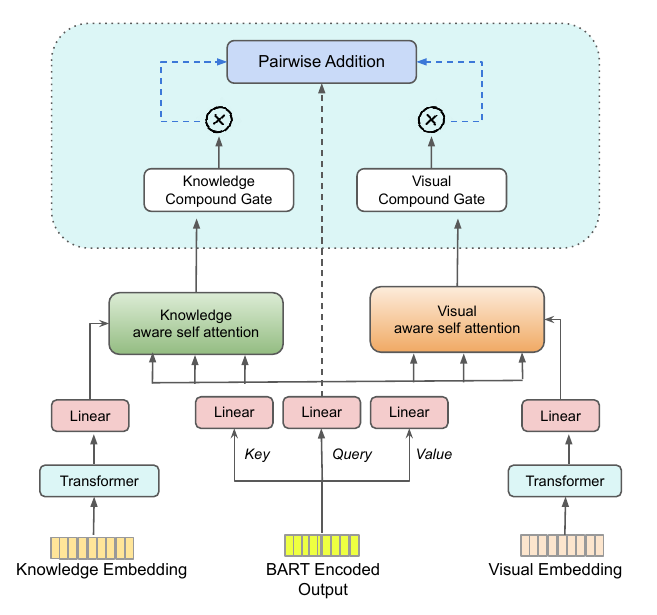}
    \vspace{-1.5em}
    \caption{Proposed modality driven knowledge infused modality fusion technique}
    \vspace{-0.9em}
    \label{IM}
\end{figure}

\begin{equation}
\small
    [Q K V] = H [W_Q, W_K, W_V]
\end{equation}

where $Q, K$, $V$ $\in \mathbb{R}^{l x d}$ are query, key, and value, respectively. Here, $l$ and $d$ denote the sequence length and the dimension of the hidden state (H). The term $W_Q, W_K$, and $W_v$ are the learnable parameters corresponding to the key vector, having the dimension of $\mathbb{R}^{d x d}$. To determine the co-relation of visuals and additional knowledge with patient-doctor interaction discourse, we generate visual and relevant knowledge conditioned key ($\hat{K}$) and value ($\hat{V}$) vectors. The attention vectors transpose the query vector (dialogue transcript) to generate a multi-modal, knowledge-aware information vector. The key and value pairs are calculated as follows: 
\begin{equation}
\small    {
\begin{bmatrix}
\hat{K} \\
\hat{V}
\end{bmatrix}} = {(1 - 
\begin{bmatrix}
{\lambda_k} \\
{\lambda_v}
\end{bmatrix}) {
\begin{bmatrix}
K \\
V
\end{bmatrix}} +
\begin{bmatrix}
{\lambda_k} \\
{\lambda_v}
\end{bmatrix}(E {
\begin{bmatrix}
{U_k} \\
{U_v}
\end{bmatrix}})}
\end{equation}
where $\lambda \in \mathbb{R}^{l x 1}$ is the learnable parameter that determines how much information from the textual modality should be retained and how much other modality information should be integrated. Here, $E$ could be evidence (visual feature) or experience (additional relevant knowledge). $U_k$ and $U_v$ are the learnable parameters. The modality controlling parameters ($\lambda$) are calculated using the gating mechanism as follows:
\begin{equation}
\small
     {
\begin{bmatrix}
\lambda_k \\
\lambda_v
\end{bmatrix}} = { \sigma ({
\begin{bmatrix}
K \\
V
\end{bmatrix}} {
\begin{bmatrix}
W_{k_1} \\
W_{v_1}
\end{bmatrix}} + E {
\begin{bmatrix}
{U_k} \\
{U_v}
\end{bmatrix}} {
\begin{bmatrix}
W_{k_2} \\
W_{v_2}
\end{bmatrix}}) }
\end{equation}
where $W_{k_1}, W_{k_2}, W_{v_1}$ and $W_{v_2}$ ($\in \mathbb{R}^{d x 1}$) are trainable weight matrices. Finally, the visual and knowledge aware attentions ($H_v$, and $H_{kn}$) and the final attended vector ($\hat{H}$) are calculated as follows:
\begin{equation}
\small
\centering
\begin{split}
& H_v  = Softmax (\frac{Q\hat{K}_{v}^{T}}{\sqrt{d_{k}}})\hat{V}_{v} \\
& H_{kn} = Softmax (\frac{Q\hat{K}_{kn}^{T}}{\sqrt{d_{kn}}})\hat{V}_{kn}
\end{split}
\end{equation}

\hspace{-0.43cm}\textbf{Fusion} In order to infuse and control the amount of information transmitted from the different modalities (visual and external knowledge), we build two compound gates: visual ($g_v$) and world-knowledge ($g_{kn}$). The context information is transmitted via the gates as follows:
\vspace{-0.3em}
\begin{equation}
\centering
\begin{split}
& g_v  = [H \oplus H_v]W_v + b_v \\
& g_{kn} = [H \oplus H_{kn}]W_{kn} + b_{kn} 
\end{split}
\end{equation}

where $\oplus$ denotes a concatenation operation. $W_v$ \& $W_{kn}$ ($\in \mathbb{R}^{2d X d}$) and $b_v$, \& $b_{kn}$ ($\in R^{d X 1} $) are parameters. The final contextualized attended vector ($\hat{H}$) is computed as Equation \ref{ff}, which is being utilized for intent identification and encoder representation. 
\begin{equation}
\small
    \hat{H} = H + g_v \odot	H_v + g_{kn} \odot	H_{kn}
    \label{ff}
 \end{equation}

\subsection{Medical Department Identification and MM-CCS Generation} 
In the healthcare system, various medical departments exist, and specialists within each department are generally more adept at comprehending relevant cases. Motivated by this understanding, we aim to leverage the knowledge of the specific medical department to enhance the generation of precise summaries for clinical conversations. Thus, we build a multi-task department identification and summary generation framework (Figure \ref{IM}) that utilizes the encoder representation to identify the medical department. The same encoded representation is then fed into the decoder to generate the summary. Note, clinical conversation summary generation is our  primary task which is being comprehended with the other task of medical department identification.

\hspace{-0.43cm} \textbf{Clinical Department} The encoder takes clinical transcript and determines attention over visual and additional relevant knowledge using the proposed contextualized M-modality fusion. The attended multi-modal encoder representation vector, $\hat{H}$ (Equation \ref{ff}), is passed to a fully connected neural network having a linear layer and nine nodes in the final layer representing different medical departments. 
 
 \hspace{-0.43cm}\textbf{Multi-modal Clinical Conversation Summary (MM-CCS)} The decoder block takes the attended multi-modal encoder representation vector ($\hat{H}$) and feeds it into the multi-head attention layer as key and value, with the query as the hidden representation of the clinical conversation text. The infused information is processed with the traditional transformer's layers of GPT-2 and computes the vocabulary's probability distribution.
 
 \hspace{-0.43cm} \textbf{Outcome Space and Loss Function} The sizes of outcome space are 9 department classes and 5845 vocabulary tokens for classification and generation tasks, respectively. We have utilized a joint categorical cross-entropy loss function, which is the sum of classification (CL) and generation (GL) tasks, i.e., $L = \alpha_1 * CL + \alpha_2 * GL  $ and $ \alpha_1(=0.2) +  \alpha_2(=0.8) =1$.

\section{Experimental Setup} We have utilized the PyTorch framework for implementing the proposed model. The proposed {\em MM-CliConSummation} generation model was trained for 30 epochs on an RTX 2080 Ti GPU, which took around 30 minutes. The proposed model has been trained, validated, and evaluated with 80\%, 5\%, and 15\% samples of the {\em ConSummation} dataset, respectively. The hyperparameter values for the model are as follows: sequence length/text (360), sequence length/visual (786), sequence length/knowledge graph (786), batch size (32), optimizer (Adam), activation function (ReLU), and learning rate (5e-05). Furthermore, the dataset and source code are available at \url{https://github.com/NLP-RL/MM-CliConSummation}.

\hspace{-0.42cm}\textbf{Baselines} We have utilized the following baselines to comprehend the efficacy and limitations of the proposed model:
\begin{itemize}
    \item \textbf{GPT-2} Generative Pre-trained Transformer-2 (GPT-2) \cite{budzianowski2019hello} is the state-of-the-art transformer-based language model trained on a humongous amount of English corpora in the self-supervised setting.  
    \item \textbf{BART}  BART \cite{lewis2019bart} is a denoising autoencoder model that is trained to reconstruct corrupted sentences.
    
    \item \textbf{T5}  T5 \cite{raffel2020exploring} is a versatile text-to-text model that combines encoder-decoder architecture with pre-training on a mixture of unsupervised and supervised tasks. 

    \item \textbf{MAF} MAF \cite{kumar2022did} is a fusion model that incorporates an additional adapter-based layer in the encoder of BART to infuse information from different modalities. 
      \item \textbf{K-CliConSummation} It is the proposed model with only knowledge infusion (w/o multimodal visual infusion).
     \item \textbf{M-CliConSummation} M-CliConSummation is the proposed model with multimodal visual (w/o knowledge infusion).
      \item \textbf{M-CliConSummation w/o fusion} It is the proposed model where different modalities, text, knowledge, and visuals are combined simply by concatenation.
       \item \textbf{KM-CliConSummation} KM-CliConSummation is the proposed framework with both multimodal visual and external knowledge infusion but without multi-tasking of department identification and summary generation.
       \vspace{-1em}
\end{itemize}

\section{Results and Discussion} 
We employed the most popular automatic evaluation metrics for summarization/text generation, namely BLEU, Rouge, and METEOR \cite{papineni2002bleu, lin2004rouge,banerjee2005meteor}, to evaluate the adequacy of summarization quality of the proposed model. The purpose of the proposed multi-task framework is to enhance the performance of the clinical dialogue summarization task by utilizing the additional task of medical department identification. Thus, the results and analysis mainly emphasized on summarization task. Based on the experiments, we report the following answers (with evidence) to our investigated research questions (RQs).

\begin{table*}[hbt!]
    \centering
    \scalebox{0.73}{
    \begin{tabular}{lccccccccccccc}
    \hline
     \textbf{Model}    & \textbf{B-1} & \textbf{B-2} & \textbf{B-3} &\textbf{B-4} & \textbf{BLEU}  & \textbf{R- 1}  & \textbf{R- 2} & \textbf{ROUGE- L}  & \textbf{METEOR} & \textbf{Jaccard Sim} & \textbf{BERT Score} & \textbf{Accuracy} & \textbf{F1-Score} \\ \hline
     GPT-2 \cite{budzianowski2019hello} & 11.65 & 5.34 & 2.22 & 0.80 & 5.00 & 21.23 & 4.64 & 20.37 & 23.41 & 0.0717 & 0.6660 & / & / \\
     BART \cite{lewis2019bart} & 9.94 & 7.18 & 5.16 & 3.72 & 6.50 & 38.37 & 18.04 & 35.50 & 18.68 & 0.1833 & 0.8378 & / & /  \\
     T5 \cite{raffel2020exploring} & 42.27 & 32.00 & 24.60 & 18.58 & 29.36 & 54.16& 31.74 & 51.24 &43.22 & 0.2582 & 0.8841 & / & / \\ 
     MAF \cite{kumar2022did} & 47.18 & 36.47 & 27.62 & 20.02 & 32.82 & 59.31& 36.93 & 49.71 & 55.10 & 0.2699 & 0.9131 & / & /  \\
     K-CliConSummation  & 47.87 & 36.57 & 27.89& 21.67&33.50 &59.45 &37.21 & 50.11 & 56.24 & 0.2724 &0.9096  & /& / \\
     M-CliConSummation &48.16 &37.74 & 28.06 & 22.46 & 34.10 & 60.10 & 37.21 & 51.16 & 55.62& 0.2766 & 0.9148 & / & / \\
     KM-CliConSummation (w/o fusion) & 35.38 & 21.40 & 12.01& 7.22& 19.00& 42.29& 18.88& 33.09 & 34.59 &  0.1536 & 0.7787 &  31.20 & 0.2201  \\
    KM-CliConSummation (w/ fusion) & 48.77 & 37.37 & 28.44 & 22.16 & 34.18 & 60.27 & 37.90 & 50.87 & 56.70 & 0.2753 & 0.9127  & / & / \\
    \textbf{MM-CliConSummation$^\$$} & \textbf{49.26} & \textbf{37.88} & \textbf{29.03} & \textbf{22.78} & \textbf{34.74} & \textbf{60.47} & \textbf{38.13} & \textbf{51.77} & \textbf{57.15} & \textbf{0.2778} & \textbf{0.9184} &  \textbf{60.68} & \textbf{0.5631}  \\
     \hline
    \end{tabular}}
    \caption{Performances of different models for multi-modal clinical conversation summary generation. Here, $\$$ indicates statistical significant findings ($p$ $<$ 0.05 at 5\% significance level)}
    \label{R1}
    \vspace{-1.5em}
\end{table*}

 \begin{table*}[hbt!]
    \centering
    \scalebox{0.73}{
    \begin{tabular}{ccccccccccccccc}
    \hline
     \textbf{Visual layer}  & \textbf{Knowledge layer} & \textbf{B-1} & \textbf{B-2} & \textbf{B-3} & \textbf{B-4} & \textbf{BLEU} &  \textbf{R-1} &  \textbf{R-2}& \textbf{ROUGE- L}  & \textbf{METEOR} & \textbf{Jaccard Sim} & \textbf{BERT Score} & \textbf{Accuracy} & \textbf{F1-Score} \\ \hline
   2 & 3 & 47.21 &35.54 & 26.86 & 20.39 & 32.50 & 59.05 & 36.19 & 49.20 & 54.65 & 0.2701 & 0.9118  &49.57 & 0.4578 \\
   2 & 4 & 47.26 & 35.23 & 26.17 & 19.69 & 32.09 & 58.96 & 35.67 &48.60 & 53.32 & 0.2636 & 0.9104 & 48.71 & 0.4469 \\
   3 & 4 & 48.49 & 36.91 & 28.06 & 21.53 & 33.75 & 60.07 & 37.09 & 50.48 & 56.02 & 0.2724 &0.9150 & 58.97 & 0.5448 \\
   3 & 2 & 48.68 & 37.06 & 28.18 & 21.62 & 33.88 & 60.01 & 37.47 & 50.06 & 56.23 & 0.2735 & 0.9146 & 52.13 & 0.5025 \\
   4 & 2 & 47.48 & 36.10 & 27.43 & 20.87 & 32.97 & 58.98 & 36.62 & 49.54 & 56.01 & 0.2683 & 0.9109 & 55.12 & 0.5325 \\
   4 & 3 & \textbf{49.26} & \textbf{37.88} & \textbf{29.03} & \textbf{22.78} & \textbf{34.74} &\textbf{60.47} & \textbf{38.13} & \textbf{51.77} & \textbf{57.15} & \textbf{0.2778}  & \textbf{0.9184} & \textbf{60.68} & \textbf{0.5631} \\
     \hline
    \end{tabular}}
    \caption{Performance of the proposed multi-modal clinical summary generation model with different modality infusion orders. There are 6 layers in the encoder (Figure \ref{PM}), and the higher layer number indicate a layer towards the end of the encoder}
    \label{R2}
    \vspace{-1.5em}
\end{table*}

 \hspace{-0.42cm} \textbf{RQ1: How does the inclusion of visual cues, such as visual signs and patients' expressions, impact the clinical patient-doctor interaction summarization task?} We experimented with different models with and without visual information for both overall summary generation and medical concern summary (MCS) generation. The obtained results are reported in Table \ref{R1} (overall summary) and Table \ref{R2} (MCS). The inclusion of visual description led to the following improvement in generation quality-- \textit{Overall summary}: BLEU (1.28 $\uparrow$), ROUGE-L (1.45 $\uparrow$), and METEOR (0.52 $\uparrow$), \textit{MCS}: BLEU (1.35 $\uparrow$), ROUGE-L (1.06 $\uparrow$), and METEOR (1.75 $\uparrow$). It also improved other evaluation metrics. The improvements by M-ConSummation for both tasks across the evaluation metrics firmly demonstrate the effectiveness of visuals in clinical conversation summary generation.

\hspace{-0.42cm}  \textbf{RQ2 (a): Can the inclusion of external knowledge offer more relevant context, thereby enhancing the quality of generated medical dialogue summaries?} Through the experiments, it became apparent that the infusion of knowledge played a pivotal role in enhancing the quality of generation, benefiting both the overall summary and medical concern summary (MCS). The knowledge infusion led to the following improvements-- \textit{Overall summary}: BLEU (0.68 $\uparrow$), ROUGE-L (0.40 $\uparrow$), and METEOR (1.14 $\uparrow$), \textit{MCS}: BLEU (0.53 $\uparrow$), ROUGE-L (0.66 $\uparrow$), and METEOR (1.05 $\uparrow$). Furthermore, we also observed that knowledge infused with simple concatenation with text/visual performs very poorly compared to one with proposed contextualized M-modality fusion. 

\begin{table}[hbt!]
    \centering
    \scalebox{0.55}{
    \begin{tabular}{lccccccc}
    \hline
     \textbf{Model}  &  \textbf{BLEU} & \textbf{R-1} & \textbf{R-2} &  \textbf{ROUGE- L}  & \textbf{METEOR} & \textbf{Jaccard Sim} & \textbf{BERT Score} \\ \hline
     GPT 2 \cite{budzianowski2019hello} & 2.45&14.1 & 3.03& 13.57 & 19.45 &0.0512 & 0.6558  \\
     BART \cite{lewis2019bart} & 23.91&46.92 & 27.11& 44.16 & 43.59 &0.2964 & 0.8784  \\
    T5 \cite{raffel2020exploring} & 26.37& 48.02 & 27.69 & 44.10 &49.28 & 0.2994 & 0.8770  \\ 
     MAF \cite{kumar2022did} & 33.83 & 61.12 & 42.23 &58.95 & 57.79 & 0.3584 &0.8778  \\ 
    K-CliConSummation  & 34.36 & 61.59 & 42.70 & 59.61 & 58.84 & 0.3630 & 0.8810 \\ 
     M-CliConSummation  & 35.18 & 62.42 & 43.18 & 60.01 & 59.54 & 0.3737 & 0.8859  \\ 
    KM-CliConSummation (w/o fusion)  &25.01 & 48.64  &29.54 &46.01 &43.36 &0.2746 & 0.8153  \\ 

     KM-CliConSummation (w/ fusion)  & 35.04 & 62.40 & 43.03 & 60.22 & 59.32 & 0.3720 & 0.8842  \\ 
     \textbf{MM-CliConSummation$^\$$}  & \textbf{35.78} & \textbf{62.83} & \textbf{43.96} & \textbf{60.90} & \textbf{59.51} & \textbf{0.3772} & \textbf{0.8862}  \\ 
     \hline
    \end{tabular}}
    \caption{Performance of the different models for medical concern summary generation.  Here, $\$$ indicates statistical significant findings ($p$ $<$ 0.05 at 5\% significance level)}
    \label{R3}
    \vspace{-2em}
\end{table}

\hspace{-0.42cm}  \textbf{RQ2 (b): Does the amalgamation technique of various modalities, namely text, visuals, and knowledge, have any influence on the quality of generated summaries?} To investigate the research question, we conducted experiments involving various techniques for integrating modalities at different model layers. The obtained results are reported in Table \ref{R1} and Table \ref{R3} (KM-CliConSummation w/o fusion and KM-CliConSummation w/ fusion). It shows that the model that incorporates modality order-driven multimodal infusion performs significantly superior to the model which simply concatenates different modalities. We anticipate that the order of modalities infusion and the distance between them is crucial to the effectiveness of combined information. Thus, we also experimented with different models having modalities infusion at different layers of the transformer (Table \ref{R2}). The findings show that the most preferred position for modality infusion in the transformer is towards the last layers (we have a total of six layers in the encoder). Moreover, knowledge should be infused before visual as knowledge is also a kind of text and thus can be merged uniformly with text.

\hspace{-0.42cm}  \textbf{RQ3: Is there a correlation between the identification of medical departments and the summarization of medical dialogues? } To investigate the research question, we experimented with a multi-task framework that identifies the medical department as well as generates a summary of a clinical conversation. The results are reported in Table \ref{R1} and Table \ref{R2}. The proposed multi-tasking framework performs superior to all baselines across various evaluation metrics for both overall summary and medical summary generation. Similar behavior is also obtained in human evaluation; the model outperformed all others, even with human perception. Note that the human evaluation conducted was performed in a blind review manner, ensuring that no information regarding the model names was provided alongside the summaries.

\hspace{-0.36cm}\textbf{Human Evaluation} We also conducted human evaluation of 100 test samples. In this assessment, two medical domain experts and one researcher (other than the authors) were employed to evaluate the generated summaries by different models (without revealing the models' names). The samples are assessed based on the following five metrics: \textit{adequacy, fluency domain relevance (DR), consistency, and informativeness (info)} on a scale of 1 (extremely poor) to 5 (idle). The obtained scores are presented in Table \ref{HE_Des}. 
\begin{table}[hbt!]
    \centering
    \scalebox{0.75}{
    \begin{tabular}{lcccccc}
    \hline
     \textbf{Model}  & \textbf{Adequacy} & \textbf{Fluency} & \textbf{DR} &  \textbf{Consistency} & \textbf{Info}  & \textbf{Avg} \\ \hline
    T5 \cite{raffel2020exploring} & 2.80 & 4.26 & 3.96 & 3.34 & 3.98 & 3.67  \\ 
     MAF \cite{kumar2022did}  & 3.24 & 4.28 &4.01  & 3.64 & 4.15 & 3.86   \\ 
     K-ConSummation  & 3.46 & 4.32 & 4.38 & 3.86 & 4.22 &4.05  \\ 
    M-ConSummation  & 3.56 & 4.31 & 4.40 & 3.82 & 4.28 & 4.07  \\ 
     KM-ConSummation  & 3.65 & 4.36 & 4.44 & 3.94 & 4.34 & 4.15  \\ 
     MM-ConSummation  & \textbf{3.88} & \textbf{4.40} & \textbf{4.56} & \textbf{4.02} & \textbf{4.42} & \textbf{4.26}  \\ 
     \hline
    \end{tabular}}
    \caption{Human evaluation of different summary generation models}
    \vspace{-2.2em}
    \label{HE_Des}
\end{table}

\hspace{-0.36cm}\textbf{Key Observations} The key observations and insights are as follows: \textbf{(i)} Gaining an understanding of one task by leveraging knowledge from a related task is consistently advantageous. The proposed model exhibits similar behavior, both in terms of modality infusion (where knowledge is infused with text first, followed by visual infusion with text, and finally combining text, attended knowledge, and visual vectors) and multi-tasking (medical department identification and summary generation). \textbf{(ii)} The model that incorporates knowledge/visual features at the initial layers of the encoder (as shown in Table \ref{R2}) exhibits subpar performance. This can be attributed to the prominent importance of the text modality in summarization, requiring some processing before merging with supplementary information from heterogeneous sources. \textbf{(iii)} We observed that the models infused with knowledge perform significantly superior for medical department and disease identification (department/disease is also included in some summaries).

\begin{figure*}[hbt!]
    \centering
    \includegraphics[scale=0.66]{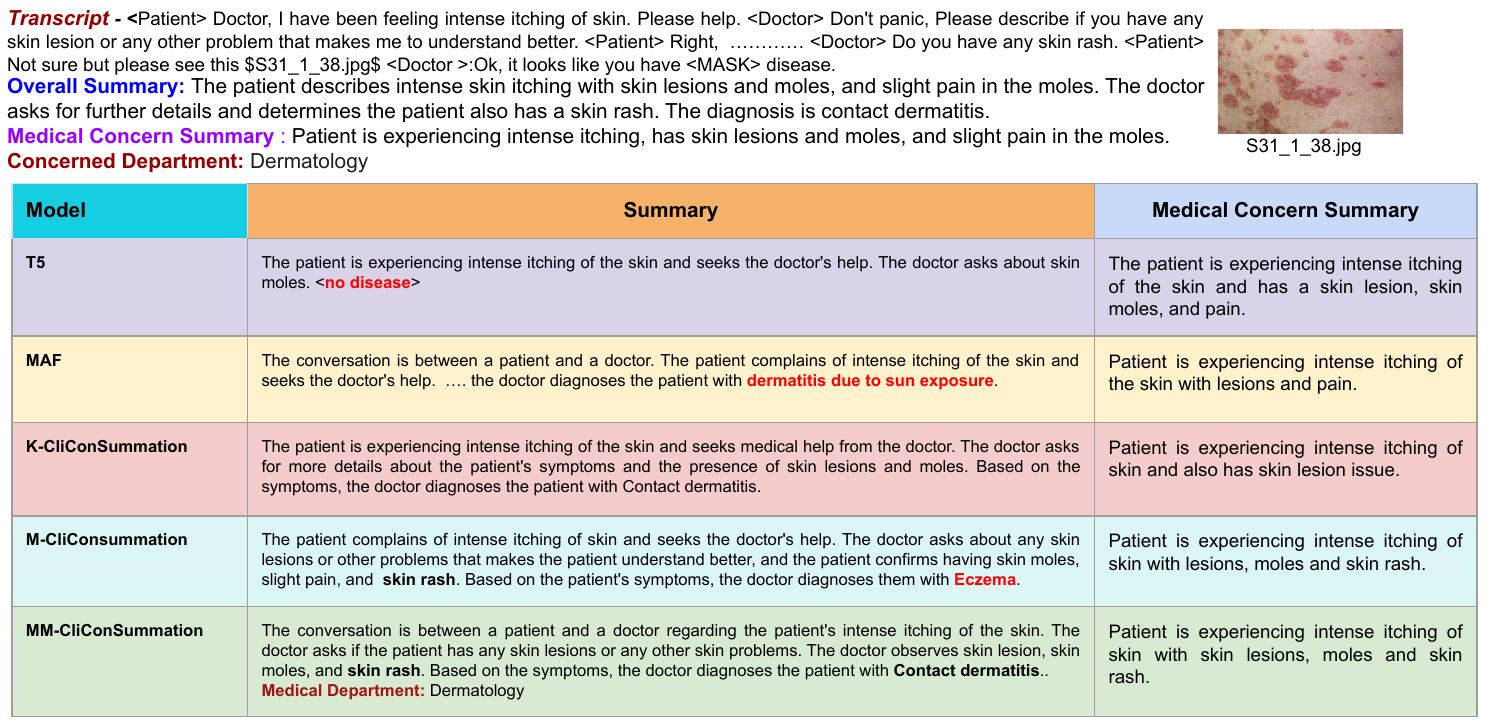}
    \vspace{-1em}
    \caption{Performance of the different baselines and proposed models for a common test case}
    \label{EA}
\end{figure*}

 \section{Analysis} We conducted a thorough qualitative analysis of the summaries generated by different baselines and our proposed models. We also performed some case studies; one such instance is illustrated in Figure \ref{EA}. The analysis leads to the following: (a) The state-of-art models (T5 and MAF--unimodal and w/o external knowledge) and baselines quite often either do not include disease or infer an incorrect disease. In the majority of cases, the predicted diseases tend to belong to the same group that encompasses the disease afflicting the patient. The behavior can be attributed to common symptoms across diseases of the same medical department. (b) In some cases, the baseline models include some most frequently occurring symptoms in the dataset despite having different contexts, such as fever and pain. We observed that the inclusion of knowledge in the proposed model has led to more factual consistency. (c) There are numerous ways to craft a summary that effectively captures the essence of a conversation. During our analysis, we encountered several instances where the BLEU score, a word-based matching metric, was relatively low. However, the generated summaries were highly relevant and had higher BERT scores, which evaluate similarity based on contextual embeddings. It is important to note that while the improvement in the BERT score margin may be less, it holds significant value since the BERT score scale differs from other metrics.

\hspace{-0.38cm}\textbf{Limitations} Despite the significant improvement demonstrated by the proposed knowledge-infused multi-modal dialogue summary model, we observed some weaknesses and limitations. (i) In a few cases, the model generates incomplete names of some long symptoms (spots for spots in vision). Also, it mises medical condition in summary in a few cases. Nevertheless the number was quite less (16/100 in human evaluation). (ii) The avg. length (in words) of gold summaries and generated summaries for test samples were 46 and 42, respectively. The proposed model tends to generate relatively small summaries, particularly for dialogues having a large number of utterances. (iii) During the process of summarizing a case for a senior doctor, junior doctors often provide comprehensive descriptions of visual symptoms, including details such as severity and the affected area. However, the {\em MM-CliConSummation} model does not delve into these specific details. Instead, it identifies symptoms from images and includes their names in the summaries. This limitation is primarily attributed to the lack of meticulousness in annotating visual symptom images.
\section{Conclusion}
In this work, we proposed the task of multi-modal clinical conversation summarization and medical concern summary generation. We curated a multimodal clinical conversation summary generation dataset, {\em MM-CliConSumm}, and annotated each conversation with two additional executive summaries, a medical concern summary, and a doctor impression. When we summarize a document, we tend to focus on some crucial evidence and part of the document relevant to the referenced document. Motivated by the observation, we propose a multi-tasking, knowledge-infused, multimodal clinical conversation summary generation, {\em MM-CliConSummation} framework. It takes clinical conversation (having both text and visual) as input and generates a precise summary, and identifies the concerned clinical department. The proposed {\em MM-CliConSummation} model extracts relevant knowledge graphs depending on dialogue context, constructs visual representation for visual reporting, and infuses them with the modality-driven contextualized fusion technique. The model identifies the concerned medical department with encoder representation, and the decoder generates a summary. With the extensive set of experiments, including human evaluation, the proposed {\em MM-CliConSummation} model demonstrated significant improvement over baselines and state-of-the-art models across all evaluation metrics for both summary and MCS generation. Summaries can be crafted using different word sequences or synonymous terms compared to the gold standard summary. Thus, relying solely on word-based matching is inadequate; semantic comprehension should also be taken into account. In the future, we aim to develop a novel loss function for summary generation that optimizes both semantic and lexical aspects.

\section{Acknowledgement}
Abhisek Tiwari expresses sincere appreciation for receiving the Prime Minister Research Fellowship (PMRF) Award from the Government of India, which provided support for conducting this research. Dr. Sriparna Saha extends heartfelt gratitude for the Young Faculty Research Fellowship (YFRF) Award, supported by the Visvesvaraya Ph.D. Scheme for Electronics and IT, Ministry of Electronics and Information Technology (MeitY), Government of India, and implemented by Digital India Corporation (formerly Media Lab Asia), which has been indispensable in facilitating this research.

\bibliographystyle{ACM-Reference-Format}
\balance
\bibliography{Ref}

\end{document}